\documentclass[letterpaper, 10 pt, conference]{ieeeconf}  % Comment this line out if you need a4paper
\IEEEoverridecommandlockouts                              % This command is only needed if you want to use the \thanks command
\usepackage[l2tabu,orthodox]{nag}
\usepackage[
backend=biber, 
bibencoding=utf8,
style=ieee, 
sorting=none, 
natbib=true, 
doi=false, 
isbn=false, 
url=false, 
eprint=false, 
maxcitenames=1, 
mincitenames=1,
maxbibnames=5,
]{biblatex}

\usepackage[pdftex,colorlinks]{hyperref}
\usepackage[printonlyused]{acronym}

\usepackage{siunitx}
\usepackage[all]{nowidow}

\usepackage[dvipsnames]{xcolor}

\usepackage{lipsum}

\usepackage[pdftex]{graphicx}

\usepackage{epstopdf}

\usepackage{import}

\graphicspath{{./latexGoodPractices/}}

\usepackage{booktabs}

\usepackage{tabu}

\usepackage{amssymb,amsfonts,amsmath,amscd}

\usepackage{bm} 
\newcommand{\bbm}{\begin{bmatrix}}
\newcommand{\ebm}{\end{bmatrix}}

\usepackage{microtype}
\usepackage{tabularx}
\usepackage{soul}

\newcommand{\FP}[1]{#1}

\newcommand{\remove}[1]{}

\title{\LARGE \textbf{Collision-Aware Traversability Analysis for Autonomous Vehicles in the Context of Agricultural Robotics}}

\graphicspath{ {./images/} }

\begin{document}

\author{Florian Philippe$^{1,3}$, Johann Laconte$^{2}$, Pierre-Jean Lapray$^{1}$, Matthias Spisser$^{3}$ and Jean-Philippe Lauffenburger$^{1}$%
\thanks{$^{1}$ Université de Haute-Alsace, IRIMAS, EA 7499, 68093, Mulhouse, France}%
\thanks{$^{2}$ Université Clermont Auvergne, INRAE, UR TSCF, 63000, Clermont-Ferrand, France}%
\thanks{$^{3}$ Technology \& Strategy Engineering SAS, 67300, Schiltigheim, France}%
}

\maketitle

\begin{abstract}
%Application agricole
% In this paper, we propose a novel perception method for safe navigation in the context of agricultural robotics.
% In the face of the current environmental challenges, robotics can play a big role in reducing the use of chemicals while answering growing needs for food production.
% However, challenges remain in the autonomy and resilience of robots in the field.
% Falling branches, tall grass, among others, are unstructured obstacles that the robot must safely overcome, understanding that such obstacle is deformable and thus safe to cross.
% As such we propose a novel traversability analysis method based on a 3D spectral map reconstructed from a LIDAR and a multispectral camera, allowing the robot to differentiate between safe and unsafe collisions.
% We provide a thorough evaluation of the multispectral metrics to detect vegetation, and use such metrics to build an augmented map.
% Using this map, we show how to compute a meaningful physics-based traversability metric, allowing the robot to safely overcome deformable obstacles depending on its weight and size.

In this paper, we introduce a novel method for safe navigation in agricultural robotics. 
As global environmental challenges intensify, robotics offers a powerful solution to reduce chemical usage while meeting the increasing demands for food production. 
However, significant challenges remain in ensuring the autonomy and resilience of robots operating in unstructured agricultural environments. 
Obstacles such as crops and tall grass, which are deformable, must be identified as safely traversable, compared to rigid obstacles.
To address this, we propose a new traversability analysis method based on a 3D spectral map reconstructed using a LIDAR and a multispectral camera. 
This approach enables the robot to distinguish between safe and unsafe collisions with deformable obstacles. 
We perform a comprehensive evaluation of multispectral metrics for vegetation detection and incorporate these metrics into an augmented environmental map. 
Utilizing this map, we compute a physics-based traversability metric that accounts for the robot's weight and size, ensuring safe navigation over deformable obstacles.

% First, a semantic segmentation is performed using vegetation indices computed on spectral data. 
% Then, based on the segmentation result, a 3D mass density map is estimated and used for navigation planning. 
% We evaluate the performance of our pipeline in terms of precision, accuracy and navigation costs, considering several vegetation indices. 
% In this context Normalized Difference Vegetation Index (NDVI) performs best as compared to other vegetation segmentation methods.

\end{abstract}

% \begin{IEEEkeywords}
% multispectral, proxi-detection, unstructured environment, UGV, obstacle detection, agriculture
% \end{IEEEkeywords}

\section{Introduction} % At the end, the reader should clearly understand the problem we wanna solve, and how we wanna solve it

%=============================================================
% 1) Robotics are cool in agriculture, but the environment contain numerous obstacles that are hard to handle, such as **vegetation** => as shown in fig1, we will collide with stuff and we need to be able to handle that, diff between colliding grass and a children

% 2) Cameras are cool, and NIR is even cooler for vegetation. But we also need lidar to get a geometric view of the scene

% 3) Thus, we have the clear contribution to fuse sensors to create a traversability map, [...]
%=============================================================

% Agriculture is hard and laborious, we need robots
% UGV are cool to help in agricultural tasks
% cite some examples
The application of robotics in agriculture has experienced a significant increase in recent years, driven by advancements in autonomous systems and the pressing need to address challenges in the sector. As noted by \citet{fountas_agricultural_2020}, Unmanned Ground Vehicles (UGV) are being developed to mitigate the labor shortages in agriculture, while simultaneously reducing the physical demands of agricultural work. \citet{lenain_agricultural_2021} have demonstrated the value of these robots for weeding tasks, replacing polluting chemical weedkillers with more environmentally-friendly mechanical methods.

\begin{figure}
	\centering
	\includegraphics[width=\columnwidth]{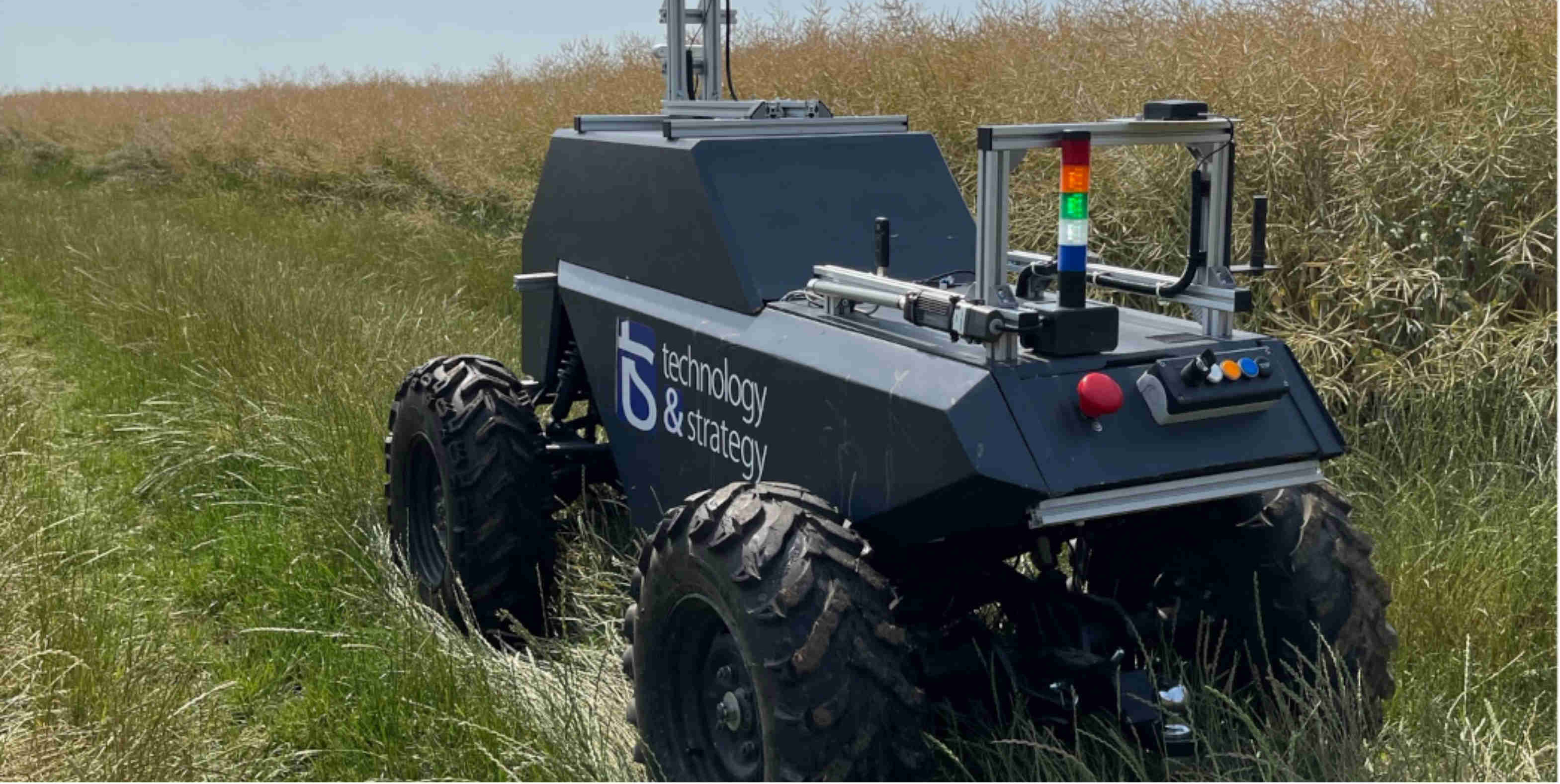}
	\caption{Autonomous plateform navigating through an agricultural field. 
	Tall grass, as well as crops, protrude into the path, forming potential obstacles the robot has to collide with.
	As such, any system working in such conditions must be able to assess the severity of each collision.}
% 	The navigation system must estimate what must be bypassed and what can be  without risk.}
	\label{fig:UGV}
\end{figure}

%the agri environment is complex and hard to sense
However, the identification of traversable elements, i.e., those which do not present a lethal threat to navigation,  is a challenging task. The main constraints in agricultural environments are the shape evolution that occur over the seasons and the presence of dense vegetation. 
Tall grass and branches may protrude in untrimmed areas, as illustrated in \autoref{fig:UGV}. 
This results in a considerable number of false detections by the perception systems, thus decreasing the UGV navigation performance and ultimately reducing the efficiency and resilience of the autonomous platform.

%We need safety guarantees while being able to undertake the task
The necessity for a high level of safety in field deployment is paramount, both for the crops themselves and for pedestrians. 
A fundamental aspect of ensuring safe and resilient navigation is the development of a comprehensive understanding of the surrounding environment. 
It is imperative that UGVs are aware of any potential danger or lack of sufficient understanding of the surrounding unstructured environment. 
Furthermore, UGVs should not halt abruptly at every blade of grass seen by the sensors.
As such, one must achieve a compromise between security and autonomy.

To address these constraints, it is crucial to identify elements with which the UGV can interact safely.
Range measurements are crucial for understanding the environment's geometry. 
Additionally, while traditional color cameras can identify scene elements, 
emerging sensors such as multi-spectral cameras provide higher spectral resolution and more detailed scene information, especially in vegetal environments.
The development of vegetation indices for detecting and monitoring plant health  \cite{abderrazak_review_1996} offers valuable new data to improve traversability analysis.

In this paper, we propose a physics-based traversability analysis, able to determine whether an element can be traversed or must be avoided by a UGV in an agricultural environment. We propose a novel method that integrates depth and spectral modalities to create a mass density-augmented map by considering the semantic nature of objects within the UGV's path, aided by precise, non-data-driven vegetation detection. 
This map is then used to evaluate potential paths in terms of the resulting loss of velocity.
The contributions presented in this article are 
\begin{enumerate}
    \item the definition of a new traversability criterion and its physical interpretation for unstructured vegetal environments using emerging sensors; and
    \item a quantitative evaluation of vegetation segmentation methods.
\end{enumerate}

%Section 2 will talk about SOTA, section 3 about the theory, ...
%\autoref{sec:related_work} of this paper will present the most advanced methods for analysing the traversability of UGV, as well as spectral analysis methods. The advantages and disadvantages will be examined. The methodology proposed will be outlined in \autoref{sec:proposed_method}. The evaluation of this approach is presented in \autoref{sec:evaluations}, with measurements taken in the field used as the basis for this assessment. In conclusion, this article will present the advantages of the proposed method and potential avenues for future improvement.

\section{Related work} \label{sec:related_work} % At the end, the reader should understand how our contributions differ from the SOTA and what we are bringing to the table

%-------Traversability analysis definition & main concepts---------
The concept of traversability analysis has been extensively reviewed in the academic literature. \citet{benrabah_review_2024} define the objectives of traversability analysis as determining whether a UGV can successfully navigate a path, or doing so while optimizing factors such as travel time or energy consumption. To achieve this goal, the surroundings are modeled during the environmental perception task. This model is used to assess navigation risks and provide crucial information for optimal and safe path planning.
\citet{beycimen_comprehensive_2023} classify traversability methods into three categories: vision-based, geometry-based, or hybrid, depending on the exteroceptive modality employed. The majority of methods rely on visual data.
Furthermore, \citet{borges_survey_2022} highlight the necessity for UGVs to have access to 6 Degrees of Freedom (DoF) pose information and large datasets to train data-driven solutions.

%-------Data-driven vs traditional methods---------
In a structured environment, \citet{laghmara_25d_2019} proposed to generate a 2.5D map by fusing data from both a LiDAR and a camera. Through the application of belief theory, it not only represents the static surroundings but also identifies dynamic objects. In contrast, unstructured environments present greater challenges.
\citet{castro_how_2023} proposed fusing height and color data with the UGV velocity to generate an environmental cost grid for offroad application. This enables more nuanced and diverse navigation behaviors compared to traditional baselines, as vehicle velocity is integrated into the decision-making process.
While most map-based approaches describe the environment with navigation costs or physical modeling (e.g., occupancy probability, slope, or ground roughness), \citet{cai_risk-aware_2022} introduced a novel criterion based on UGV physics. To evaluate traversability, the environment is modeled in terms of the appropriate speeds for UGVs to navigate specific cells, employing an AI-driven solution.
%AI training on the field has limits
Both methods from \citet{castro_how_2023} and \citet{cai_risk-aware_2022} require a setup phase to train the AI model for risk estimation in navigating the environment. However, assessing the risk of traversing impassable areas—such as mud, which poses a sinking hazard, or crops that must be avoided—remains challenging without a highly realistic simulator. The proposed method does not need to be train\FP{ed} in-situ.
%So does have AI solution for safety
%In a review of data-driven terrain segmentation, \citet{kabir_terrain_2025} highlight the complexities of natural terrains, which include grass, gravel, rocks and water, requiring specific detection and segmentation algorithms. Variability in terrain appearance due to changes in lighting, weather and seasonal conditions increases ambiguity, complicating accurate classification.
Moreover, the success of data-driven segmentation methods strongly depends on the quality, quantity, and diversity of training data \cite{liu_computing_2021}, making this approach unsuitable for security and environmental protection applications.
%Non-data-driven solution are proposed
%In contrast, \citet{ollis_structural_2003} proposed a traversability method that does not rely on AI, instead using heuristics to estimate the semantic characteristics of each cell in a 2.5D grid. A risk value is assigned to each semantic class based on the UGV's capabilities. While this heuristic approach can handle specific situations, it poses challenges in terms of generalization.
%What we propose
To address AI's safety limitations, this paper proposes the use of a model-based semantic segmentation approach.

%-------Traversability criteria fusion---------
In unstructured environments, relying on a single criterion may be insufficient.
\citet{leininger_gaussian_2024} proposed a mapless method for traversability analysis. Geometric features (slope, step height, flatness) are extracted from a Gaussian Process before to be fused in a local traversability map.
\citet{fan_step_2021} also proposed a multi-criteria traversability approach. Factors such as slope, ground roughness, and steps are evaluated to create a multi-layered, risk-aware cost grid. The effectiveness of their approach has been demonstrated through field trials conducted in magma tubes and an abandoned metro line.
However, one may wonder whether the criteria previously described are able to discern passable elements. In the proposed method, the impact of collisions is evaluated using semantic information.

%-------Vegetation segmentation---------
%Transition
\remove{Indeed, the dense vegetation, common during spring and summer in agricultural environments, can hinder the effectiveness of traversability analysis. Vegetation obstructs visual features and introduces additional geometric information that must be disregarded. Being able to detect this vegetation would allow for the integration of its crossability into the traversability analysis.
Most of the UGVs visual data currently comes from grayscale or color cameras. However, these sensors have inherent limitations, as the light spectrum extends well beyond visible light, and the three-channel input is insufficient to capture its full complexity. Spectral analysis offers a more comprehensive approach to vision, and multispectral cameras help to address this shortcoming.}
\FP{Dense vegetation, common in agricultural environments during spring and summer, can obstruct traversability analysis by blocking visual features and introducing unnecessary geometric information. Detecting vegetation would enable the integration of its crossability into the analysis. While most UGVs rely on grayscale or color cameras, these sensors are limited by the narrow visible light spectrum and poor spectral resolution (up to 3 channels). Multispectral cameras enhance vision by providing higher spectral resolution and capturing data beyond the visible spectrum, thereby addressing this limitation.}
The use of spectral distances, as discussed by \citet{richards_image_2006} for satellite image classification, has been widely studied, with applications in detecting crops, concrete, and water. These methods assess how closely spectral measurements align with a reference profile.
%Vegetation indices introduction
Additionally, green vegetation absorbs red light while reflecting near-infrared (NIR) radiation \cite{gitelson_relationships_2003}. As noted by \citet{abderrazak_review_1996}, vegetation indices are employed to monitor these spectral bands, providing insights into plant health.
Based on this assumption about the environment, \FP{the proposed traversability method incorporates vegetation detection using spectral images.}
%Vegetation index based perception solution
In the work of \citet{santos_segmentation_2021}, vegetation indices were utilized for vegetation detection, specifically comparing indices based on visible light for segmentation tasks. The Modified Green Red Vegetation Index (MGRVI) demonstrated the best performance in segmenting green plants. Furthermore, Otsu and K-Means binarization algorithms were evaluated, with K-Means delivering superior results in most scenarios.
Vegetation segmentation using data-driven methods has also been explored. \citet{sa_weednet_2018} introduced WeedNet, a segmentation model that employs a CNN with visible and NIR inputs. The network classifies \texttt{weeds}, \texttt{crops}, and \texttt{background}, with applications in weed control.
For navigation tasks, \citet{kulic_deep_2017} developed a segmentation network to detect trails in unstructured environments using a multispectral camera to enhance navigation.

%-------Vegetation segmentation linked to the environment ~closest papers---------
%proxi-detection
Few studies have been conducted on the application of spectral measurements for proximity detection. \citet{zou_multi-spectrum_2017} proposed fusing vegetation indices with LiDAR depth data for obstacle detection. Superpixel generation on the \FP{Normalized Difference Vegetation Index (}ndvi\FP{)} image facilitates the detection of non-plant elements.
A Support Vector Machine (SVM) classifier is then applied to categorize obstacles based on various geometric and visual features. 
However, this method is measurement-based. In a map-based approach, false detections could be filtered out by leveraging prior knowledge, due to most of the environment being static over time.
To build an environmental model, \citet{clamens_real-time_2021} combined multispectral and LiDAR measurements into a 3D map of the environment, augmented with the \remove{Normalized Difference Vegetation Index (}ndvi\remove{)}. These 3D maps are subsequently used to monitor the health of plantations, as well as for fruit detection and counting operations using data-driven methods.
In the proposed method, a similar map is generated to conduct traversability analysis based on the semantic segmentation of vegetation within the environment.
Plant detection is employed to estimate the crossability of map elements.
%Transition 
\autoref{sec:proposed_method} introduces a traversability index that accounts for the semantic characteristics of scene elements to estimate their traversability. This method addresses the issue of false detections caused by dense vegetation in agricultural environments.

\section{Map-based Traversability Estimation} \label{sec:proposed_method} %Proposed method

%Data flow for traversability analysis schema
\begin{figure*}
	\centering
	\includegraphics[width=\textwidth]{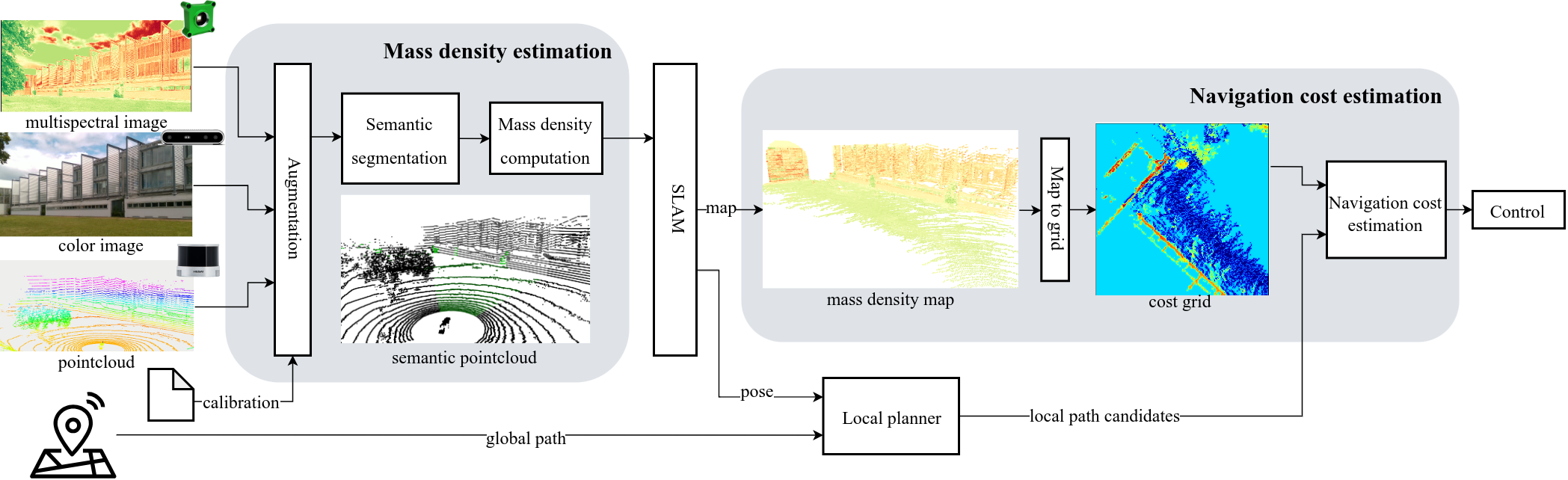}
	\caption{Traversability analysis flow: Multispectral images are fused with LiDAR data to produce an augmented point cloud, which serves as the basis for semantic segmentation. Mass density is calculated for each depth measurement, updating the 3D environmental map. This is then converted into a 2D traversability grid. Ultimately, potential paths are assessed, and the safest one is chosen.}
	\label{fig:tta_flow}
\end{figure*}

%Intro section
The objective of the proposed method is to estimate a path navigation cost by computing the potential loss of velocity that would be experienced by the UGV traversing it. 
\autoref{fig:tta_flow} illustrates the traversability analysis pipeline.
Initially, the spectral measurements obtained from the camera are projected into the point cloud of the LiDAR. 
A semantic segmentation of the environment is conducted using spectral measurements. 
A mass density is estimated for each LiDAR point in the camera's field of view, on the basis of the probabilities of belonging to a given semantic class. 
The augmented point cloud is then filtered and stored within a 3D map, before being projected onto a 2D grid in order to estimate the traversability of the terrain. 
%In accordance with the mass and velocity of the UGV, along with its local trajectory candidates, the loss of velocity for a given trajectory is estimated.
The loss of velocity for a given trajectory is estimated.
Local paths leading to a larger velocity loss are rejected, and thus a safe trajectory is selected. 
% It is next transmitted to the UGV control solutions.

\subsection{Augmentation} %Preliminaries
The first step is to process the raw sensor data to derive a spectrally-enhanced depth measurement. After performing the necessary calibrations, the LiDAR data is projected onto the camera measurements. We refer to the outcome of this procedure as the \textit{spectrally augmented point cloud}.
%3D spectral augmented map
%Calibration process
In order to fuse spectral and spatial information on a 3D map, three calibrations are required:
\paragraph{Intrinsic calibration} estimates the distortion coefficients of the optics, the focal lengths and the optical center coordinates forming intrinsic matrix $\bm{K} \in \mathbb{R}^{3 \times 4}$ of each camera. 
\paragraph{Extrinsic calibration} identifies the relative pose between each sensor and provides a transformation matrix defined by $\bm{T} \in SE(3)$.
\paragraph{Spectral calibration} \citet{sattar_snapshot_2022} proposed a calibration procedure to establish a connection between reflectance, i.e., the physical material properties that determine how light is reflected, and the light intensity sensed by the camera. This step leads to \textit{multispectral image} illustrated in \autoref{fig:tta_flow}. The conversion is achieved through the use of a linear system model. The spectral reflectance vector, denoted as $\bm{r} \in \mathbb{R}^{m}$, is estimated based on the corresponding spectral intensity vector $\bm{i} \in \mathbb{R}^{n}$, using the spectral calibration matrix $\bm{M}\in \mathbb{R}^{m\times n}$, for each pixel of the camera image. The reflectance vector is defined as
\begin{equation}
   \bm{r} = \bm{M}\bm{i}.
   \label{equ:reflectance_estimation_function}
\end{equation}

From this, the lidar point cloud is projected on the image frame using
\begin{equation}
    \bm{p}_\text{image} = \bm{K}\bm{T}\bm{p}_\text{lidar},
  \label{equ:spectral_projection}
\end{equation}
where $\bm{p}_\text{lidar} \in \mathbb{R}^{4}$ and $\bm{p}_\text{image} \in \mathbb{R}^{3}$ denotes the depth data measured by the LiDAR in their respective frames.
As such, the LiDAR point cloud is augmented with spectral information from the camera, that will be used in the following to estimate a mass density map.
This process is illustrated in  \autoref{fig:tta_flow}, where it is denoted as \textit{\FP{A}ugmentation}.
\FP{The semantics are assessed using the spectrally augmented point cloud. This step is presented in \autoref{fig:tta_flow} as \textit{Semantic segmentation}. It is \FP{both described and evaluated} in the \autoref{sec:semantic_seg}.} 
% Index definition
%   Criteria
%Fusion short explanation
% Then, the spectral data is used to perform semantic segmentation of the scene.

\subsection{Mass density map generation} \label{subsec:mass}
\remove{t}\FP{T}he aim of this section is to estimate the mass density of the environment.
Therefore, we define the \textit{mass density augmented point cloud} as a set of vectors of the form $[x,y,z,d_m]^T$, where $d_m\in\mathbb{R}_{\geq 0}$ denotes the associated mass density.
This step is denoted as \textit{Mass density computation} in \autoref{fig:tta_flow}.
For this, we propose a data-driven approach: each of the $n$ semantic classes, $c_i$, is associated with a reference mass density $d_m(c_i)$.
Furthermore, to address the inherent uncertainties, the mass density is modeled as a random variable. % on the probability $p(c_i)$ of belonging to each class. 
The mass density is computed as the expected value alongside the possible obstacles classes probabilities $p(c_i)$, as
\begin{equation}
\begin{aligned}
    \mathbb{E}[d_m] &= \sum_{i=1}^{n}d_m(c_i) \cdot p(c_i | s) \\
                    % &=  \frac{\sum d_m(c_i)p(\bm{y}|c_i)p(c_i)}{p(\bm{y})} \\
                    &=\frac{\sum_i d_m(c_i)p(s|c_i)}{\sum_i p(s|c_i)}, 
\end{aligned}
   \label{equ:density_estimation_function}
\end{equation}
where $s$ is the semantic measurement, and $p(c_i | s)$ are the confidence on the measurements. Such values can be either input in the framework, or learned from an annotated dataset.

%SLAM
Once the 4D point cloud is processed, it is transformed into a 3D map to be used for traversability analysis.
% A 3D map is initially generated, and sensor poses are estimated over time using a LiDAR graph SLAM, as introduced by \citet{koide_portable_2019}. 
The 3D map is generated by aggregating the 4D measurements into a voxel grid, converting them into the world frame. The sensor poses is determined using the SLAM LiDAR solution from \citet{koide_portable_2019}.
Next, the 3D map is converted into a 2D grid to assess navigation costs.
This process is depicted in \autoref{fig:tta_flow} as \textit{Map to grid}. 
This is done by flattening the map, where ground points are filtered using the RANSAC algorithm \cite{fischler_random_1981}, along with points above the UGV's height. 
The mass density grid is initialized with the UGV mass value and is later filled with the maximum mass density value from the corresponding Z-axis column.
As such, a 2D grid map with mass density information is generated, that is used in the next section to estimate the loss of velocity a robot would undergo given a path in the environment.
% The traversability grid generated thus far represents the environment around the UGV in terms of mass density. 

\subsection{Velocity loss evaluation}
% Traversability analysis on augmented maps
% how to compute the lost velocity
%\JL{add intro sentence: Using the mass density map, we derive a physics-based formula for traversability, relying on the loss of velocity.}
Using the mass density grid, we derive a physics-based formulation for traversability, relying on the loss of velocity due to navigation in nonempty space.
This step is designated as \textit{Navigation costs estimation} in \autoref{fig:tta_flow}.
Assuming inelastic collisions, the final velocity $v_R^f$ of the UGV is computed as
\begin{equation}
   v_R^f = \frac{m_R}{m_R+m_i} v_R \Leftrightarrow v_R^f = \alpha v_R,
   \label{equ:inelastic_collision}
\end{equation} 
where $v_R$ is the initial velocity of the UGV, $m_R$ the mass of the robot, and $m_i$ the mass of the $i$th obstacle.
% whereof an element of initial velocity $v_R$ and mass $m_R$ after colliding with a static element $i$ of mass $m_i$ using the inelastic collision theory.
% As such, $\alpha$ denotes the velocity loss coefficient. For $N$ collisions of masses $\{m_i\}_{1,...,N}$, the coefficient is
% \begin{equation}
%    \alpha = \prod_{i=1}^{N} \frac{m_R}{m_R+m_i}.
%    \label{equ:velocity_lost_coeff}
% \end{equation}

From this, we model the environment as a collection of infinitesimal particles of area $\Delta a\to 0$.
Given a path $\mathcal{P}\subset\mathbb{R}^2$ crossing a total area $A$, the velocity coefficient $\alpha$ of the robot is given by
\begin{equation}
   \alpha = \prod_{i=1}^N \frac{m_R}{m_R+d_m(i)\Delta a},
   \label{equ:velocity_lost_coeff}
\end{equation}
assuming colliding with $N$ particles of size $\Delta a$ such that $N\Delta a=A$. 
Note that free space is modeled as a particle of null mass density. 
From this, with the particle area $\Delta a$ tending towards zero, the equation becomes
\begin{equation}
\begin{aligned}
   \alpha &= \lim\limits_{\Delta a \to 0} \prod_{i=1}^{A/\Delta a} \frac{m_R}{m_R+d_m(i)\Delta a} \\
          &= \exp{\left(- \frac{1}{m_R}\int_\mathcal{P} d_m(a)  da\right)},
   \end{aligned}
   \label{equ:velocity_lost_coeff_2}
\end{equation}
where $d_m(a)$ denoted the local mass density at the given position in the environment.
As such, for infinitesimal particles, \autoref{equ:velocity_lost_coeff_3} compute the ratio of lost velocity is undergoing the path $\mathcal{P}$.
In the case of a grid map in which the mass density is constant inside each cell, the equation can be simplified to  
\begin{equation}
\begin{aligned}
   \alpha &= \exp{\left(- \frac{1}{m_R}\sum_{c_i\in\mathcal{P}} d_m(i) a_c\right)},
   \end{aligned}
   \label{equ:velocity_lost_coeff_3}
\end{equation}
where $a_c$ is the area of a grid cell.
One can note that assuming infinitesimal collisions result in a lower bound on the loss of velocity, meaning that we will always overestimate the risk for one path.

To summarize, the environment is modeled through the multimodal fusion of LiDAR and cameras, as described in \autoref{equ:spectral_projection}. As the UGV navigates, the grid is continuously updated to represent the mass density of surrounding elements using \autoref{equ:density_estimation_function}. This grid is subsequently used to evaluate routes candidates using \autoref{equ:velocity_lost_coeff_3}, allowing the selection of the safest path.
This method enables the environment to be physically characterized by the expected loss of velocity by crossing a specific region.
As such, the ratio of velocity $\alpha$ will be used as the navigation cost.
In the following section, we will qualitatively assess this approach and quantitatively evaluate a non-data-driven vegetation semantic segmentation solution.

\section{Evaluations} \label{sec:evaluations}
In this section, we outline the setup employed for evaluating the proposed method. Next, we assess the performance of the semantic segmentation indices, followed by an offline test of our traversability analysis solution using recorded data.
\subsection{Experimental Setup}
%Hardware presentation
% A series of measurements was conducted at INRAE's test sites. 
%\JL{Using fig3 picture, describe the environment used in the experiments. We don't care it was at INRAE}
A teleoperated platform was equipped with a Hesai Pandar XT-32 LiDAR, a Silios CMS-V VNIR multispectral camera, and an Intel Realsense D415 stereoscopic camera. The VNIR camera provided 8 spectral measurements from \SI{550}{\nm} to \SI{830}{\nm}, in addition to a panchromatic (PAN) measurement. The stereoscopic camera provided 3 channels of RGB measurements.
Data are collected using this mobile measurement bench, enabling us to evaluate our traversability analysis solution offline. As illustrated in \autoref{fig:3d_map}, one of the environments explored was a park with tall grass, untrimmed trees and winding paths. Buildings and static pedestrians are present throughout the navigation. The weather was mild during the recordings.

% Vegetation segmentation evaluation
\subsection{Semantic segmentation performance} \label{sec:semantic_seg}
%Present compared methods

\begin{figure}
	\centering
	\includegraphics[width=\columnwidth]{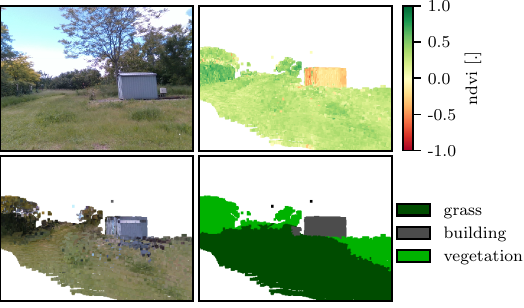}
	\caption{Augmented 3D maps of a park environment, consisting of tall grass, bushes, trees and a small shack. Top Left: Color camera's image from the scene; Top Right: ndvi colorized 3D map; Bottom Left: visible light colorized 3D map; Bottom Right: Manually annotated 3D map}
	\label{fig:3d_map}
\end{figure}

In \autoref{subsec:mass}, we presented a generic way to estimate the mass density of the environment using semantic classes.
As such, the quality of this estimate rely heavily on the level of accuracy on the segmentation.
The following section presents a thorough quantitative evaluation of vegetation indices and spectral distance performances, highlighting their performance to differentiate vegetation from other objects.
The evaluation is conducted using a manually annotated 3D dataset, where spectral information was fused with a 3d point cloud. 
Each point on the map is labeled with one of the following classes: \texttt{Grass, Track, Vegetation, Building, Pedestrian, Obstacle} or \texttt{Other}.
An example of the annotation and spectral measurements is depicted in \autoref{fig:3d_map}. In total, \SI{6228592}{} points were annotated for a total covered area of \SI{2515}{\m\squared}, consisting of off-road environments.

% Show an augmented map

The present study focuses on the vegetation detection:
a macro-class designated \texttt{Plants} is defined, encompassing both the \texttt{Vegetation} and \texttt{Grass} classes, while all other classes are included under the $\neg \texttt{Plants}$ macro-class.

\begin{figure}
	\centering
	\includegraphics[width=\columnwidth]{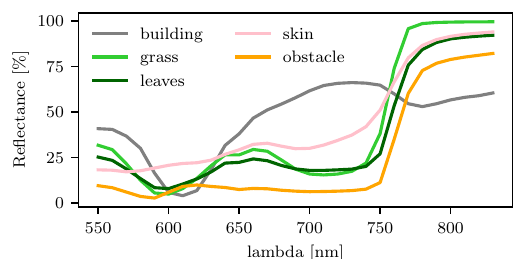}
	\caption{Reflectance profiles of several elements of agricultural environment}
	\label{fig:ref_plot}
\end{figure}

In the following, we evaluate the most popular metrics to detect and quantify the vegetation. Namely,  the Modified Green-Red Vegetation Index (mgrv), Green Leaf Index (gli), Modified Photochemical Reflectance Index (mpr), Red-Green-Blue Vegetation Index (rgbvi), Excess of Green (exg), Excess of Red (exr), Vegetative (veg), Normalized Difference Vegetation Index (ndvi), and Enhanced Vegetation Index (evi) \cite{abderrazak_review_1996}, are evaluated.
Additionally, the spectral distances, such as Euclidean Distance (ed), Bray-Curtis Distance (bc), and Spectral Angle (sa) \cite{richards_image_2006, kruse_spectral_1993}
are also compared.
The reference reflectance profiles used for the computation of spectral distances were extracted from the annotated spectral maps for each class by averaging the spectral measurements. These profiles consist of 29 wavelengths, ranging from \SI{550}{\nm} to \SI{830}{\nm}, and are presented in \autoref{fig:ref_plot}.

%Reflectance profile used for spectral distance process

The segmentation procedure is described as follows: the vegetation index and spectral distance are applied to the spectral augmented maps. Subsequently, a binarization step is conducted using the Otsu algorithm \cite{otsu_threshold_1979} to segment vegetation with a dynamic threshold.
% Evaluation metrics
The  \textit{Intersection over Union (IoU), Precision (Prec.), Accuracy (Acc.), Recall (Rec.), F1 score, Specificity (Spec.)} and \textit{computation duration} $\Delta t$ are compared for each vegetation segmentation method in \autoref{tab:vegetation_segmentation_performance}. 
%Hardware
An AMD Ryzen 7 5000 series CPU is used for the computational duration measures.

\begin{table}[!ht]
    \caption{Benchmarking \FP{of} model-based vegetation indices for vegetation segmentation.}
    \label{tab:vegetation_segmentation_performance}
    \begin{tabularx}{\columnwidth}{c|XXXXXXc}
       \toprule
       Index    & IoU               & Prec.         & Rec.          & Acc.          & F1            & Spec.         & $\Delta t$ [ms] \\
       \midrule
       %Vegetation Index
        mgrv    & 0.54              & 0.94          & 0.56          & 0.62          & 0.70          & 0.84          & 169.5             \\
        gli     & 0.42              & 0.97          & 0.43          & 0.53          & 0.59          & 0.94          & 107.5             \\
        mpri    & 0.49              & 0.94          & 0.51          & 0.58          & 0.66          & 0.88          & \textbf{60.3}     \\
        rgbvi   & 0.68              & 0.87          & 0.76          & 0.71          & 0.81          & 0.53          & 148.3             \\
        exg     & 0.56              & 0.98          & 0.57          & 0.64          & 0.72          & 0.94          & 82.9              \\
        exr     & 0.33              & 0.78          & 0.37          & 0.41          & 0.50          & 0.57          & 64.1              \\
        exgr    & 0.64              & 0.89          & 0.70          & 0.69          & 0.78          & 0.63          & 138.8             \\
        veg     & 0.80              & 0.84          & \textbf{0.94} & 0.81          & 0.89          & 0.27          & 194.8             \\
        evi     & 0.56              & 0.99          & 0.56          & 0.64          & 0.72          & \textbf{0.98} & 104.0             \\
        ndvi    & \textbf{0.91}    & \textbf{0.99}  & 0.92          & \textbf{0.93} & \textbf{0.95} & 0.94          & 67.8              \\
       \midrule
       %Spectral distances
        sa      & 0.69              & 0.79          & 0.84          & 0.69          & 0.81          & 0.05          & 883.7             \\
        bc      & 0.66              & 0.95          & 0.69          & 0.72          & 0.80          & 0.85          & 1589.4            \\
        ed      & 0.90              & 0.96          & 0.93          & 0.91          & \textbf{0.95} & 0.84          & 430.1             \\
       \bottomrule
    \end{tabularx}
\end{table}
As such, the most reliable method for identifying vegetation in proximate detection applications is the ndvi.
However, its response is dependent on the wavelength of only two specific frequencies, since the ndvi is computed as
\begin{equation}
   \text{ndvi} = \frac{i_{810}-i_{650}}{i_{810}+i_{650}},
   \label{equ:ndvi}
\end{equation}
where the intensity of light at a specific wavelength $\lambda$ nm is denoted as $i_\lambda$. False positives are likely in the case of certain elements with similar reflectance in the red and NIR bands (e.g., plastics and textiles). Pedestrians represent a small portion of the dataset, so false detections related to their clothing have a negligible effect on the overall performance results presented.
Spectral distance, such as the spectral angle, measures the consistency of a measurement with respect to a reference profile over a much higher resolution.
At the expense of processing time, this allows for greater robustness to local similarities in the analyzed light spectrum.
As such, when deadline with more complex environments where there is a thinner granularity than only differentiating between vegetation and non-vegetation, these metrics would prove themselves more robust. 
Such metrics will be investigated in future works.

Given the results in \autoref{tab:vegetation_segmentation_performance}, the normalized ndvi index is used to generate belonging probabilities of \texttt{Plants} class in \autoref{equ:density_estimation_function}.

\subsection{Navigation costs estimation analysis}
In this section, we present an application of our method.
For simplicity, we focus on only two classes that are \texttt{Plants} and $\neg \texttt{Plants}$, where future works will focus on using more classes for safe navigation.
% Mass index evaluation
It is imperative that the $\neg \texttt{Plants}$ class is not navigable; therefore, the concrete mass density is set for this class \cite{noauthor_nf_2014}. 
In contrast, the \texttt{Plants} class permits navigation. 
The respective mass densities $d_m(\texttt{Plants})$ and $d_m(\neg \texttt{Plants})$ are set to \SI{20}{\kilogram\per\m\squared} and \SI{2400}{\kilogram\per\m\squared}. 
The UGV mass is set to \SI{250}{\kilogram}.
The cost grid is next extracted from the mass density-augmented map, as described in \autoref{sec:proposed_method}.

\begin{figure}
	\centering
	\includegraphics[width=\columnwidth]{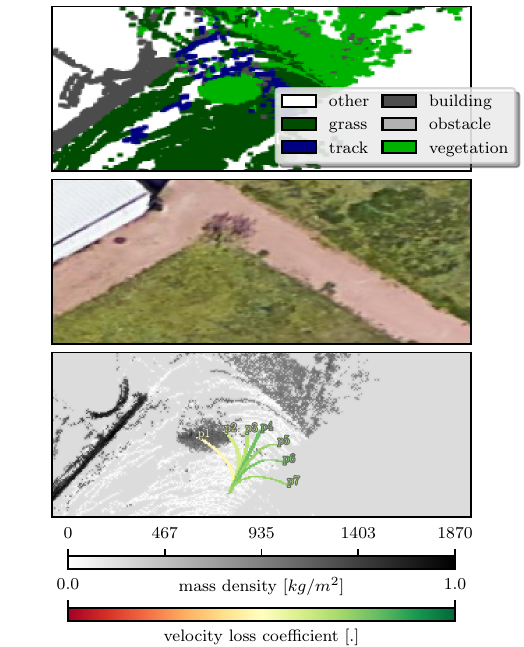}
	\caption{Local planning candidates evaluated based on the mass density grid. Semantic grid (top) and satellite view (middle) illustrate the environmental configuration during the run (middle). The local paths over the cost grid (bottom) illustrate navigation costs of each candidate. A coefficient of $1$ means no loss of velocity, thus a safe path, whereas a coefficient of $0$ means a collision resulting in the total stop of the robot.}
	\label{fig:path_costs}
\end{figure}

\autoref{fig:path_costs} illustrates an example of navigation. In this scenario, a path-following solution presents the UGV with a series of potential local paths, each of which offers a different route to reach the desired objective. These routes are then evaluated using the navigation cost function, which is applied to the mass density grid. The environment is characterized by the presence of high grass, dense vegetation (e.g., bushes, trees) and buildings, as illustrated on the semantic grid. The UGV's local planner generates seven candidate paths (color pixels) over the cost grid (gray pixels).

The path p1 guides the vehicle through a grove of trees, with a navigation cost of  ($\alpha_1=0.4$). This results in a significant reduction in speed due to the collision with the substantial vegetation. Path p7 directs the UGV to an uncharted region.
The navigation cost is ($\alpha_7=0.7$) due to the grid initialization and the unknown nature of the objects it contains. Ultimately, path p4 traverses exclusively through regions with null density mass. 
As shown in the label grid, the path is comprised solely of grass, which is a component of the ground. The navigation cost of p4 is ($\alpha_4=0.9$), indicating that the UGV will experience a minor loss of speed when navigating this route. The local planner selected this path over the other candidates.
%Paragraph dedicated to compare with SOTA
In this context, a non-semantic-aware method, as described in \citet{leininger_gaussian_2024}, would treat vegetation as an impassable step for the vehicle.

%\JL{add a discussion on the method: is it extendable to many classes? is the behavior intuitive? Could we use other metrics (energy based?)}
This method is based on semantic understanding of the environment. It can be employed in diverse environments settings and extended with different semantic classes. Furthermore, it can be adapted to any UGV's size, from a lawnmower to a tractor, by setting the appropriate mass value $m_R$. Finally, the loss velocity coefficient $\alpha$ used to evaluate a candidate path can be derived to estimate more complex metrics, such as the loss of kinetic energy over a path.

%Paragraph dedicated to estimate RT capabilities
\FP{The grid-based approach for mass density estimation is inherently highly parallelizable and the ndvi requires relatively low computation. This potentially makes the solution suitable for real-time applications.}

%Predicted costs vs. measure costs

\section{Conclusion} \label{seq:conclusion}
In this paper, we presented a novel traversability analysis method specifically designed for Unmanned Ground Vehicle (UGV) navigation in agricultural environments. The proposed approach constructs a representation of the environment that incorporates both impassable obstacles and crossable elements within the scene.

A quantitative evaluation of algorithms for vegetation segmentation was conducted, demonstrating that the use of the Normalized Difference Vegetation Index (ndvi) yields the most effective results in terms of detection accuracy and computational efficiency.

We provided an example of the use of our framework for traversability analysis, showing that the robot is able to differentiate between dangerous and safe collisions.
The approach can be adapted to various UGV models, allowing for flexibility in how obstacles are defined based on specific vehicle requirements.

For future work, a quantitative evaluation of the collision-aware navigation method will be conducted, linking the proposed solution to robot control. Additional semantic classes could be explored for density map estimation, utilizing either the spectral angle distance, or a data-driven approach with appropriate safeguards. 
Furthermore, integrating the mass index into a multi-layer grid would be advantageous, as it would enable the inclusion of additional features to enhance the traversability estimation. These features could include ground slope, height map, and other relevant characteristics.
%Furthermore, this approach does not take into account the presence of mobile obstacles. The addition of a multiple object tracker to the solution would enable the removal of measurements from moving objects for the purpose of updating the map. 
%Ultimately, this methodology can be integrated with a local planner for autonomous field trials, thereby validating its efficacy.

\section*{Acknowledgment}
\remove{We would like to express our gratitude to the \textit{Association Nationale de la Recherche et de la Technologie} (ANRT) and \textit{Technology \& Strategy Engineering SAS} for providing financial support for this research project, as well as to \textit{GdR IASIS} for their contribution towards the interlaboratory mobility, which was instrumental in enabling us to undertake this work.}
\FP{We gratefully acknowledge the financial support from the \textit{Association Nationale de la Recherche et de la Technologie} (ANRT) and \textit{Technology \& Strategy Engineering SAS}, as well as the contribution of \textit{GdR IASIS} towards interlaboratory mobility, which was essential for conducting this research.}

\printbibliography

\end{document}